\documentclass{article}

\usepackage{microtype}
\usepackage{graphicx}
\usepackage{booktabs} 

\usepackage{caption}
\usepackage{subcaption}
\newsavebox{\imagebox} 

\usepackage{hyperref}

\usepackage[accepted]{icml2023_arxiv}

\usepackage{amsmath}
\usepackage{amssymb}
\usepackage{mathtools}
\usepackage{amsthm}

\usepackage[capitalize,noabbrev]{cleveref}

\theoremstyle{plain}

\theoremstyle{definition}

\theoremstyle{remark}

\usepackage[textsize=tiny]{todonotes}

\icmltitlerunning{Diffusion Models for High-Resolution Solar Forecasts}
\begin{document}
\twocolumn[
\icmltitle{Diffusion Models for High-Resolution Solar Forecasts}



\icmlsetsymbol{equal}{*}

\begin{icmlauthorlist}
\icmlauthor{Yusuke Hatanaka}{ics,nimbus}
\icmlauthor{Yannik Glaser}{ics}
\icmlauthor{Geoff Galgon}{nimbus}
\icmlauthor{Giuseppe Torri}{atmo,nimbus}
\icmlauthor{Peter Sadowski}{ics,nimbus}
\end{icmlauthorlist}
\icmlaffiliation{ics}{Information and Computer Sciences, University of Hawaii Manoa, Honolulu, Hawaii, USA}
\icmlaffiliation{nimbus}{Nimbus AI LLC, Honolulu, Hawaii, USA}
\icmlaffiliation{atmo}{Atmospheric Sciences, University of Hawaii Manoa, Honolulu, Hawaii, USA}
\icmlcorrespondingauthor{Peter Sadowski}{peter.sadowski@hawaii.edu}

\icmlkeywords{diffusion models, weather, climate}

\vskip 0.3in
]


\printAffiliationsAndNotice{}  

\begin{abstract}
Forecasting future weather and climate is inherently difficult. Machine learning offers new approaches to increase the accuracy and computational efficiency of forecasts, but current methods are unable to accurately model uncertainty in high-dimensional predictions. Score-based diffusion models offer a new approach to modeling probability distributions over many dependent variables, and in this work, we demonstrate how they provide probabilistic forecasts of weather and climate variables at unprecedented resolution, speed, and accuracy. We apply the technique to day-ahead solar irradiance forecasts by generating many samples from a diffusion model trained to super-resolve coarse-resolution numerical weather predictions to high-resolution weather satellite observations. 
\end{abstract}

\section{Introduction}

Current methods for forecasting weather and climate rely on numerical simulations of the Earth's atmosphere. These simulations characterize the atmosphere in terms of coarse three-dimensional (3D) grid cells and use physical models to describe the time evolution of atmospheric variables such as temperature, pressure, and water vapor. The only obvious way to improve these models is to perform the simulations at higher spatiotemporal resolution at an ever-increasing computational cost. Forecast uncertainties are estimated by running a simulation multiple times with perturbed inputs to generate an ensemble of potential outcomes. An ensemble typically consists of tens of samples --- enough to estimate the variance of a forecast, but not enough to estimate the risk of rare events. For example, the National Oceanic and Atmospheric Administration's Global Ensemble Forecast System \cite{NOAAGEFS} comprises 21 samples.

Machine learning can improve these forecast models in two ways: (1) by improving the forecast models directly by correcting limitations of the physics-based models, and (2) via super-resolution of coarse-grained forecast model outputs, an approach known as \textit{downscaling} in the geophysical sciences. Large quantities of data exist for training these models, from both observations and simulations, with video-like spatiotemporal structures that make them amenable to modeling with deep convolutional neural networks. Recent examples from the literature include precipitation forecasting over local regions using a mix of radar and satellite data~\cite{sonderby2020metnet, ravuri2021skilful}, and forecasting global atmospheric variables using reanalysis data~\cite{pathak2022fourcastnet, bi2022pangu, lam2022graphcast}.

\begin{figure}[ht!]
\hspace{0.5cm}
\includegraphics[width=\linewidth]{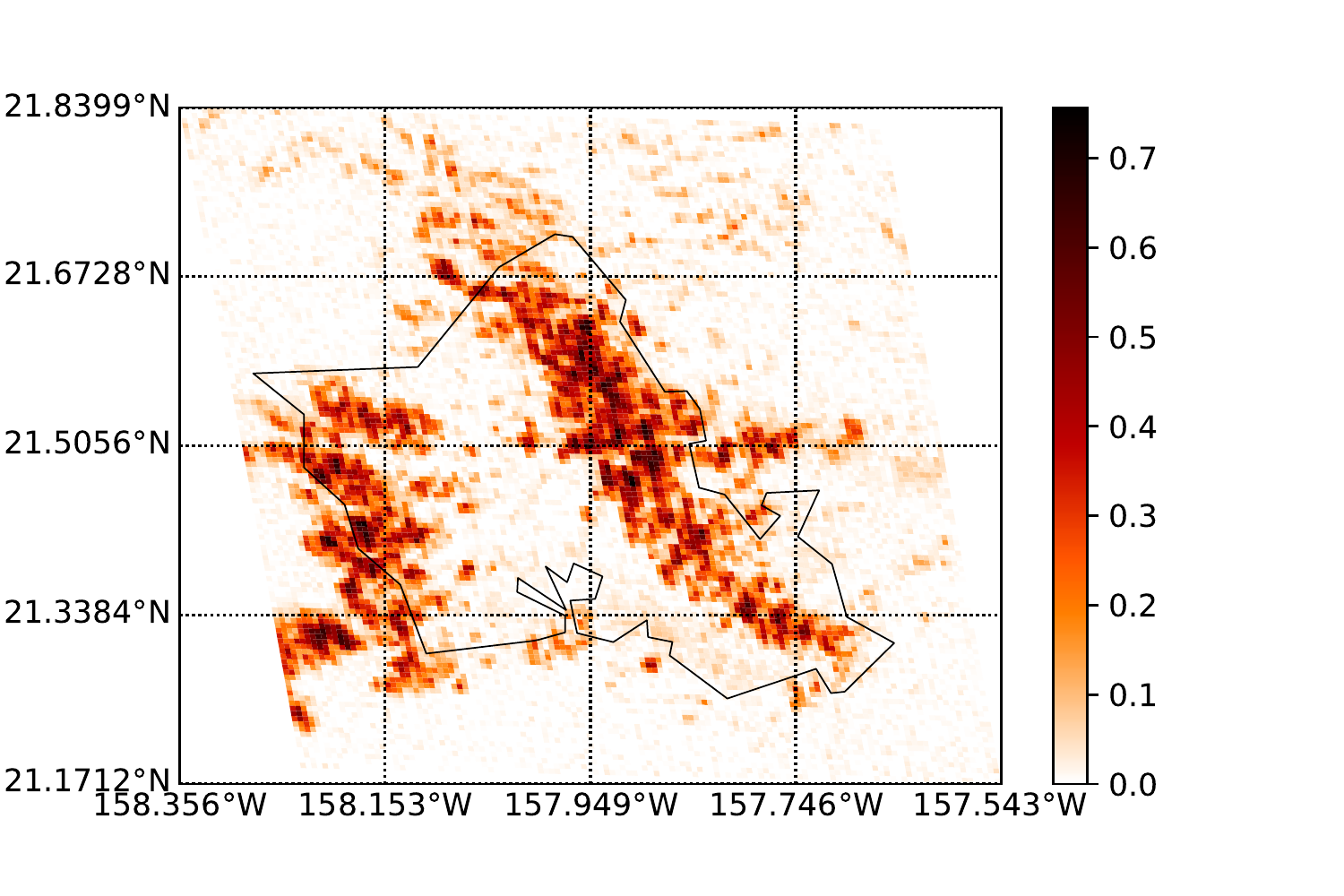}
\caption{Instantaneous cloud cover over the Hawaiian island of Oahu, sampled from a score-based diffusion model trained on satellite data with 0.5~km resolution. The high cloud density on the windward (east) side of the Koolau mountain range (center), is characteristic of mountainous tropical islands at these latitudes.}
\label{fig:oahu}
\end{figure}

However, the high degree of aleatoric uncertainty, or inherent unpredictability, presents a challenge for machine learning models. Small differences in the initial atmospheric conditions can cause large differences in outcomes, so limited-precision computational models need a way of characterizing uncertainty over a high-dimensional joint distribution of dependent variables. The most common modeling approach is to assume the predicted variables are each conditionally independent given the input of initial conditions, but this approach can fail in modeling the risks of important weather events resulting in part from the variables' joint relationships.
For example, a model might accurately predict the rainfall at each individual location in a region reasonably well (i.e. the marginal distributions) but fail to quantify the risk of flooding which requires modeling the \textit{joint} distribution of rainfall over a watershed integrated over time. 
Some work has attempted to model this uncertainty using variational autoencoders (VAEs)~\cite{kingma2013auto} or generative adversarial networks (GANs)~\cite{ravuri2021skilful}, but in this work, we make the case for using diffusion models to model uncertainty in weather forecasts. 

Score-based diffusion models have emerged as a remarkably effective approach to approximating distributions over natural images. Prominent examples include Dall-E 2~\cite{ramesh2021zero}, GLIDE~\cite{nichol2021glide}, and Imagen~\cite{saharia2022photorealistic}, which generate realistic images from text captions by combining large language models with diffusion models that sample (conditionally) from a high-dimensional image distribution. In this work, score-based diffusion models are trained on satellite images to perform super-resolution of numerical weather forecasts. A probabilistic forecast is generated by rapidly sampling from the conditional distribution defined by the diffusion model. Experimental results on day-ahead solar irradiance forecasting for the Hawaiian island of Oahu demonstrate increased accuracy along with quantified uncertainty. Our results suggest that this approach could be useful for a wide range of applications in the geophysical sciences, such as predicting precipitation and evapotranspiration or the effects of climate change on local weather patterns.

\section{Related Work}

\begin{figure*}
\centering
\includegraphics[width=\linewidth]{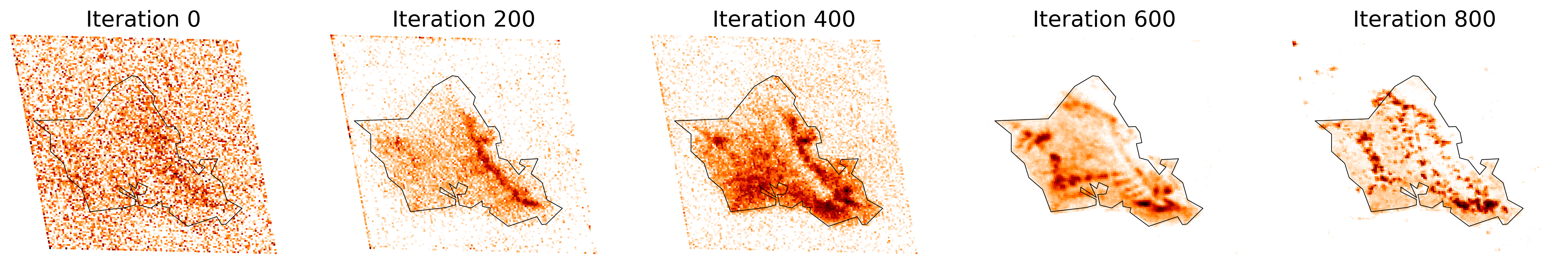}
\caption{A score-based diffusion model samples from a high-dimensional data distribution by first sampling from a reference distribution, then solving the reverse-time ODE defined by Eq.~\ref{eq:ODE} to obtain a sample from the learned distribution. In this example, a $128\times128$ pixel ``noise'' image is sampled from a multivariate Gaussian (Iteration 0) and iteratively refined by a neural network to obtain a sample of a realistic-looking weather satellite radiance image of Oahu (Iteration 800).}
\label{fig:reverse_diffusion}
\end{figure*}

\subsection{Score-Based Diffusion Models}

Diffusion models are \textit{generative} in that they describe a probability distribution $p(\mathbf{x})$, where $\mathbf{x}$ can be a high-dimensional random variable, for example, an image or the state of a physical system. In general, learning a probability distribution over a high-dimensional space from data is extremely challenging because the number of required samples grows exponentially with dimensionality. However, learning is possible when the data lives on low-dimensional manifolds, such as natural images and other data with spatial and temporal structures.

The last decade has seen a number of innovative deep-learning approaches to address the problem of learning a generative model in high dimensions. These include autoregressive models~\cite{uria2016neural}, GANs, VAEs, and flow-based models~\cite{rezende2015variational,dinh2016density}, all of which have been used heavily in scientific applications. Score-based diffusion models are a special case of \textit{reversible generative models}, which parameterize a one-to-one mapping between a known distribution and the data distribution. Previous reversible generative models such as NICE~\cite{dinh2014nice}, Real NVP~\cite{dinh2016density}, and GLOW~\cite{kingma2018glow} can be trained by directly maximizing the data likelihood, but require significant computation for each parameter training update and scale poorly to large datasets. \textit{Score-based} diffusion models provide an efficient stochastic gradient learning algorithm by approximating the score function (the gradient of the log probability density $\nabla_\mathbf{x} \log{p(\mathbf{x})}$) rather than the probability density function $p(\mathbf{x})$. The score function is easier to learn because local updates can be made during training without the need to ensure the probability density function integrates to one.

A score-based diffusion model consists of a neural network representing a time-dependent function with the same input and output dimensions, $\mathbf{s}_\theta(\mathbf{x}, t):(\mathbb{R}^D, [0,T]) \rightarrow \mathbb{R}^D$, parameterized by $\theta$. For image data, this is typically implemented as a U-Net to model the local structure in images. The data-generating process can be defined by sampling $\mathbf{x}_{T}$ from a standard multivariate normal distribution and then solving the following ordinary differential equation in the reverse time direction to generate a sample at $t=0$, $\mathbf{x}_0$:
\begin{align}
    \frac{d\mathbf{x}}{dt} &= - \mathbf{s}_\theta(\mathbf{x}, t) a(t) + \mathbf{b}(\mathbf{x}, t), \label{eq:ODE}
\end{align}

where $a(t)$ and $\mathbf{b}(\mathbf{x}, t)$ are pre-specified \textit{drift} and \textit{diffusion} coefficients that determine the shape of the prior. Because the data-generating process is reversible, we can compute the likelihood $p_\theta(\mathbf{x})$ of any data point $\mathbf{x}$. But rather than maximizing the training data likelihood directly, it is more efficient to optimize a score-based objective such as the ``denoising'' objective suggested by \citet{song2021maximum}:
\begin{align}
    \mathcal{J} &=  \mathbb{E}_\mathbf{x_0} \mathbb{E}_{t} \mathbb{E}_\mathbf{x_t} \left[ a(t) \lVert \mathbf{s}_\theta(\mathbf{x}_t, t) - \nabla_{x_t} \log p(\mathbf{x}_t | \mathbf{x}_0) \rVert_2^2 \right]
\end{align}

where the outer expectation is over the training data set, the second expectation is over time $t \sim \mathcal{U}(0,T)$, and the inner expectation is over a distribution of sample corruptions in which the training sample $\mathbf{x}_0$ is propagated through a stochastic noising process for time $t$ (in practice, this is just $\mathbf{x}_0$ plus some Gaussian noise which increases with $t$). The objective is efficient to optimize using stochastic gradient descent. At each iteration, samples are drawn from each of the three distributions, and parameters are updated with gradient descent. The result is a simple, intuitive, training algorithm: Gaussian noise is added to data samples and the parameterized model is trained to denoise the samples. This objective has been shown to upper bound the negative log-likelihood of the training data~\cite{song2021maximum} and thus the model learns to approximate a high-dimensional joint distribution. In this work, we generate many samples from such a model to provide a probabilistic forecast.

\subsection{Solar Irradiance Forecasting}

Solar irradiance is a particularly interesting forecasting application since better forecasting can reduce the risks to electricity grid stability that are associated with increasing grid penetration of non-dispatchable solar photovoltaics (PV). 
The last few years have seen a number of applications of computer vision techniques to solar irradiance forecasting, including optical flow models~\cite{wood2012cloud} or more recently deep learning~\cite{zhang2018deep,sun2018convolutional,hart2020nowcasting,berthomier2020cloud,kellerhals2022cloud}, including GANs~\cite{nie2021sky}. However, none of these approaches provide fully-probabilistic predictions that have the ability to estimate the risk of many PV sites being cloudy at once; this work is the first to propose a method for estimating this risk in a tractable way. 

We evaluate our approach on the Hawaiian island of Oahu (population 1,000,000 and the site of ICML 2023), where the problems and opportunities of renewable energy are amplified. The isolation of each Hawaiian island's electrical grid and a dependence on imported petroleum result in high electricity prices, with Hawaii consumers paying \$0.30/kWh in 2021, twice that of customers in California (geopolitical events in May 2022 caused this to increase 34\% to \$0.42/kWh~\cite{bls_solar_2022}). On the other hand, Hawaii's low latitude makes solar an attractive alternative energy source, and Honolulu has more PV production per capita than any other city in the United States, with PV systems installed on nearly a third of single-family residential homes~\cite{penn_hit_2022, heco_sustainability_report_2022}. This results in large spikes in energy demand when clouds pass over solar generation sites and dense population areas. Predicting these events could enable grid operators to mitigate their impact through demand response strategies such as starting new generators, charging batteries, and turning off water heaters and HVAC systems.

\section{Methods}

\subsection{Model}
\begin{figure*}
\centering
\includegraphics[width=\linewidth]{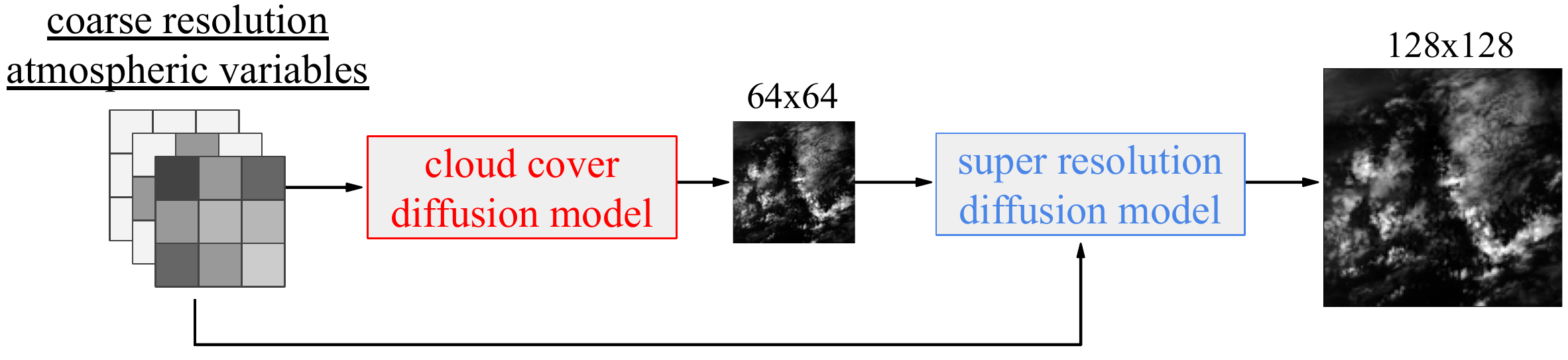}
\caption{Coarse resolution output from a numerical weather prediction model is converted to a high resolution cloud cover image using two diffusion models. The first model takes the atmospheric variables and generates a $64\times64$ pixel cloud cover image. The second model is conditioned on the same atmospheric variables, and increases the resolution to $128\times128$ pixels.}
\label{fig:pipeline}
\end{figure*}

In experiments we use two cascaded diffusion models~\cite{ho2021cascaded}, where the first model generates $64\times64$ pixel images, then a second model performs super-resolution to produce $128\times128$ pixel images  (Fig. \ref{fig:pipeline}). Both diffusion models are conditioned on the same coarse-resolution atmospheric variables output by numerical simulations. At training time, the model is conditioned on atmospheric variables from ERA5 reanalysis data. The trained model is then evaluated for two different applications: (1) \textit{historical} cloud cover prediction by conditioning on held-out reanalysis data; (2) \textit{future} cloud cover prediction by conditioning on forecasts from GFS. GFS and ERA5 use the same underlying physics models, so the same atmospheric variables are available at approximately the same resolution.

\begin{figure}
\centering
\includegraphics[width=\linewidth]{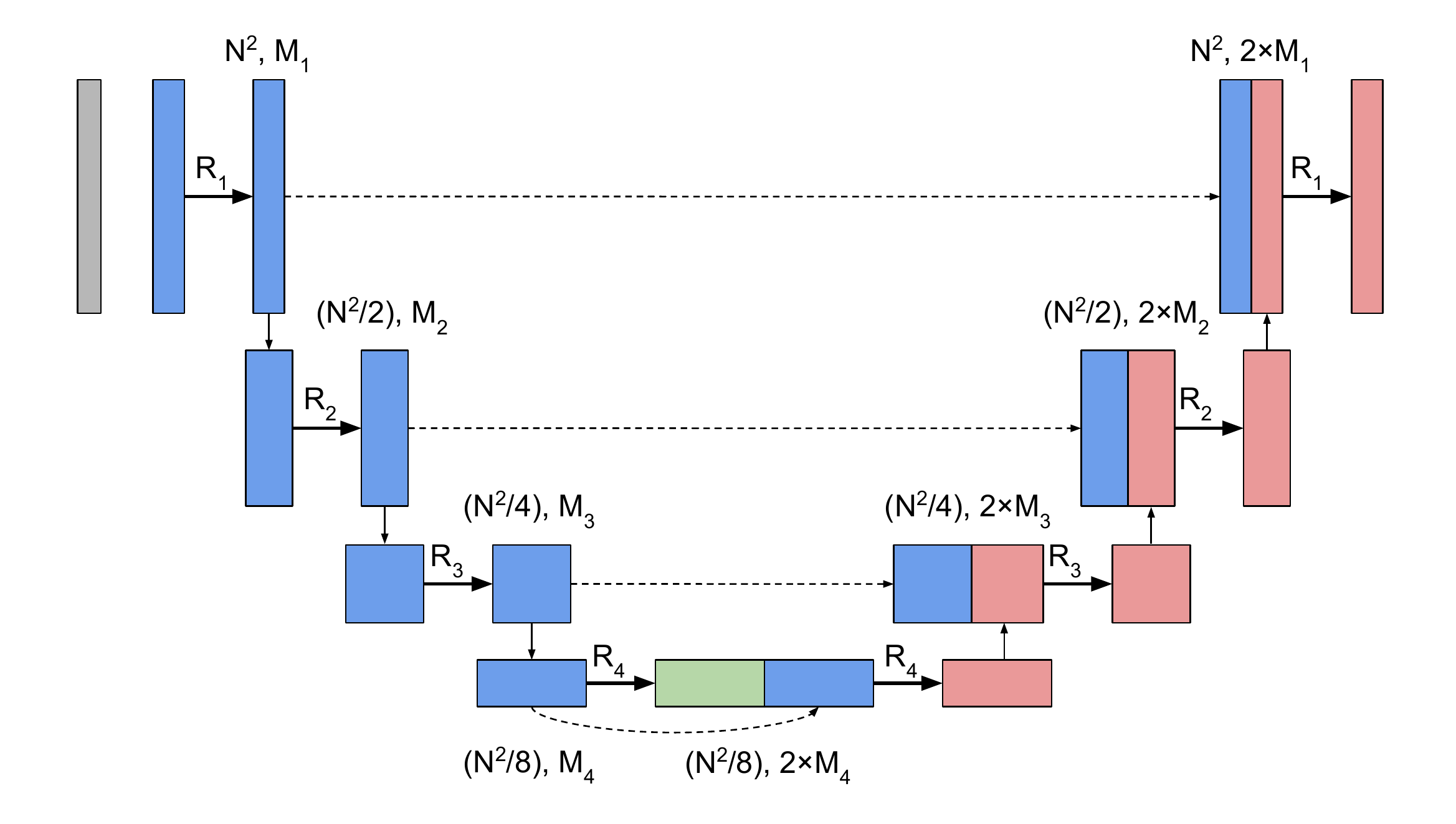}
\caption{Both diffusion models use a U-net architecture. Vertical lines are downsampling/ upsampling steps, and dotted horizontal lines are concatenations. $R_n$ and $M_n$ denote the number of residual blocks and the channel-dimension multiplier, respectively. $N$ is the dimension of the input image, i.e., $N=64$ for the first diffusion model, and $N=128$ for the super-resolution diffusion model.}
\label{fig:architecture}
\end{figure}

A U-Net architecture~\cite{ronneberger2015u} is used for each diffusion model (Figure~\ref{fig:architecture}). It utilizes the skip connections to preserve the context at different spacial levels, while the network progressively encodes the image to capture information on wider range. The specific U-net configurations largely follow details provided in \citet{ho2021cascaded} for the 64$\times$64$\rightarrow$128$\times$128 ImageNet model. Briefly, each downsampling and upsampling block adjusts the spatial dimensionality by a factor of 2 while the channel count is modified by a multiplier $M_n$. The atmospheric conditioning information is injected as a flat vector into each block in the network. 
Our full cascaded diffusion model implementation is based on the publicly available code by~\citet{imagen-pytorch}.

\subsection{Data}

Three types of data are used for experiments. The model is trained to predict cloud cover images derived from GOES satellite data from reanalysis data. Then this trained model is used to super-resolve GFS forecasts. Model training was performed on data from January 2019 -- June 2021, and all model evaluations were performed on data from July 2021 -- June 2022. 

\subsubsection{GOES satellite data}

The National Oceanographic and Atmospheric Administration's GOES-17 (GOES-West) satellite provides high-resolution atmospheric measurements over the Pacific Ocean in near real-time with the explicit goal of improving weather forecasting capabilities. Its Advanced Baseline Imager conducts full-disk observations once every 10 minutes, measuring 16 spectral bands from visible to long-wave infrared. We use Band 2 ($0.60 - 0.68$~{\textmu}m), which has 0.5~km spatial resolution and matches the peak absorption range of crystalline silicon photovoltaic (PV) cells. Data from 2019--2022 was downloaded from the Google Cloud Platform.

From the GOES-17 measurements of solar radiance, we estimated total cloud cover for each pixel using a flat-fielding algorithm. Working with this representation of the data has three advantages. First, total cloud cover does not drastically change throughout the day, as is the case with solar radiance or ground irradiance. Second, total cloud cover can be combined with data-driven clear-sky radiation models such as \citet{perez1990modeling} to predict solar irradiance on the ground. This approach is more accurate than the non-data-driven estimates of global horizontal solar irradiance provided by numerical weather prediction. Third, total cloud cover is a variable provided by both GFS and ERA5, making it easy to compare our predictions with those predictions.

For each timestamp, the zenith solar angle was calculated relative to the center of the island of Oahu and removed if its value is more than 80 degrees, in order to avoid noisy data points. This resulted in a dataset of 51,830 training data and 22,807 test data. See Fig. \ref{fig:gt_samples} for some of the random data examples from the training dataset. Both the training and the test set include hours between 6 am and 6 pm, although the range varies depending on the season.

\subsubsection{ERA5 reanalysis data}

Historical reanalysis data was taken from the European Center for Medium-range Weather Forecasts (ECMWF) Reanalysis v5 (ERA5)~\cite{Hersback2020}. Reanalysis data are produced through a blend of observational data and short-range weather forecasts that get assimilated and the result is incorporated into one regularly spaced grid. While most reanalysis products are available at spatial and temporal resolutions much coarser than typical short-range weather forecasts, they can nevertheless provide an overall understanding of past weather conditions, and they can be used to initialize numerical simulations. ERA5 is available at a grid resolution of 31 km and a temporal resolution of 1 hour, and they cover a period from 1950 to the present. 

The diffusion model is conditioned on five atmospheric variables over a $3\times3$ grid covering the island of Oahu, for a total of 45 values. The five atmospheric variables describe the cloud structure and moisture flux: (1--3) cloud cover at different levels of the atmosphere (low, medium, and high cloud cover), (4) the total cloud cover (which is simply the sum of low, medium, and high), and (5) the vertically-integrated eastward water vapor flux. For each of the variables, we extract the $3\times3$ region that covers the island of Oahu.

\subsubsection{GFS historical forecasts}

Historical GFS forecasts were downloaded from the NCAR Research Data Archive. We used the same-day forecast issued at 12:00 UTC (2:00 am Hawaii Standard Time) for the instantaneous total cloud cover at 21:00 UTC (11:00 am HST). This is a scenario highly relevant to decisions on whether to discharge energy from batteries during the morning energy demand peak depending on subsequent expected PV generation at midday. These forecasts have a coarse resolution of $0.25^{\circ}$ ($\sim$27~km in Hawaii), and we use the same 5 atmospheric features over the same $3 \times 3$ grid covering Oahu for input to the diffusion model in the forecasting experiments.

\section{Results}

\begin{table*}[h!]
\caption{Test set prediction error in RMSE for the diffusion model, persistence model, ERA5, and GFS. All RMSEs were computed using the satellite-derived cloud cover as ground truth. In the historical prediction scenario, the results are averaged over all daylight hours, while the future prediction (forecast) scenario includes only examples for 11 am local time. The difference between the diffusion model and persistence model is statistically significant in both the historical and future prediction scenarios, with the p-value computed from a paired t-test between the means. A more detailed breakdown of these results is available in the Appendix.}
\vskip 0.15in
\begin{center}
\begin{small}
\begin{sc}
\begin{tabular}{lcccccc}
\toprule
Prediction & Weather Model & Diffusion & Persistence & ERA5 & GFS & p-value\\
\midrule
Historical & ERA5  & 0.207$\pm$ 0.075 & 0.223$\pm$ 0.090 & 0.362$\pm$ 0.190 & --- & $<5\times 10^{-4}$\\
Future (11am) & GFS   & 0.198$\pm$ 0.075 & 0.202$\pm$ 0.089 & --- & 0.358$\pm$ 0.229 & 0.008\\
\bottomrule
\end{tabular}
\end{sc}
\end{small}
\end{center}
\vskip -0.1in
\label{tab:comparison_table}
\end{table*}

\subsection{ERA5 super-resolution}
On the held out test set, we evaluated predictions from the diffusion model using the total cloud cover derived from satellite data as ground-truth. Diffusion model predictions were produced by sampling many cloud cover images for a given timestep (N=45), conditioning on the atmospheric variables provided by ERA5, and taking the mean image. The root mean squared error (RMSE) was used to measure prediction error. 

\begin{figure*}[h!]
\centering
\includegraphics[width=\linewidth]{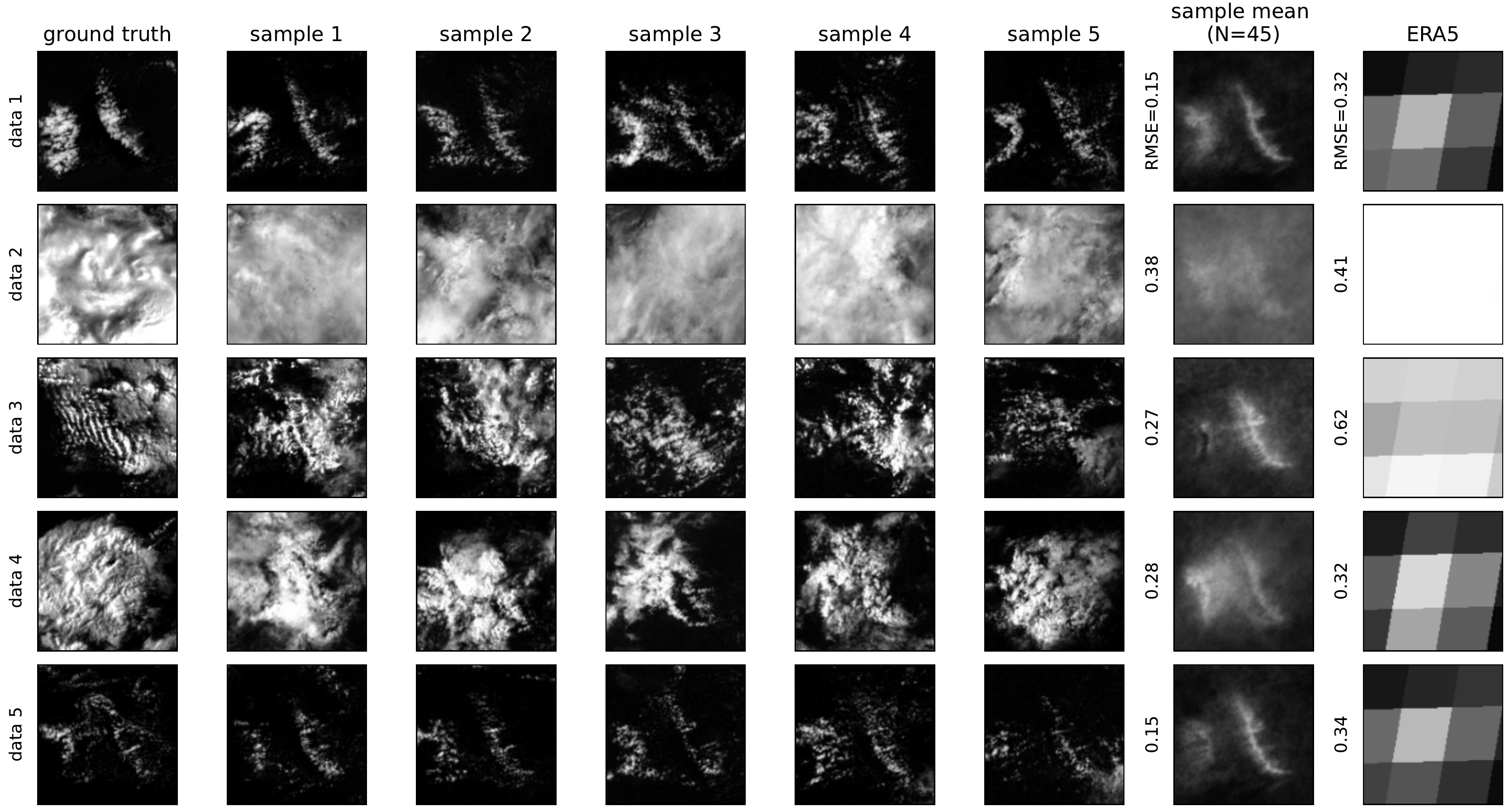}
\caption{Cloud coverage over Oahu (ground truth) is predicted by sampling repeatedly from a diffusion model (samples 1-5) and computing the mean (sample mean). This prediction has lower pixel-wise RMSE than the coarse-resolution weather data (ERA5) that the diffusion model is conditioned on because the diffusion model has learned about local weather patterns from high-resolution satellite data. We show only the top-five best images in terms of RMSE in this plot (samples 1-5 for each row), and provide the entirety of the generated images in appendix Fig.\ref{fig:all_samples_era5} for some of these data points.}
\label{fig:samples_era5}
\end{figure*}

The diffusion model prediction error was compared to two baseline prediction methods: (1) the coarse resolution total cloud cover from ERA5, and (2) a high resolution mean over all satellite-derived cloud cover images in the training set from the same time of day; we refer to this as the \textit{persistence} model. The diffusion model takes the former as an input, but can improve upon this prediction by incorporating learned patterns that account for local topography. This can be clearly seen in Figure~\ref{fig:samples_era5}, where the samples tend to have clouds accumulated at the mountainous regions of the island. The persistence model also includes these features, but is unable to account for any variations in weather. The diffusion model accounts for both of these factors and has lower RMSE (Table~\ref{tab:comparison_table}). A paired t-test between the RMSEs of the diffusion model and those of the persistence model was conducted because the mean RMSE values were close; the differences were statistically significant ($p<.01$) in both experiments, since the test dataset is a large sample.

\subsection{GFS forecasts}

\begin{figure}[ht!]
\hspace{0.5cm}
\includegraphics[scale=0.45]{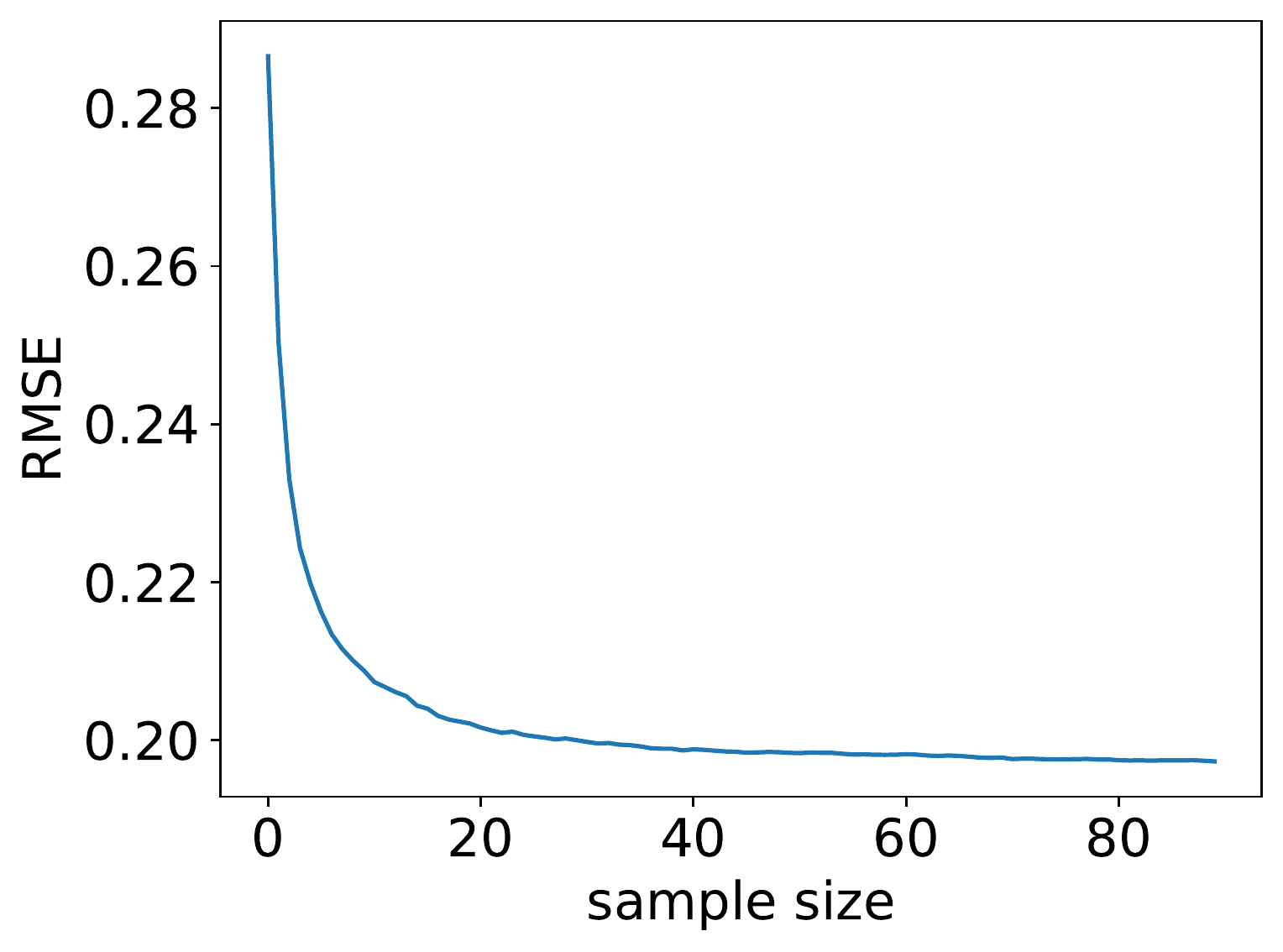}
\caption{Prediction error in RMSE vs. the number of samples (N) from the diffusion model. The prediction error is an average over the held-out test set.}
\label{fig:number_of_samples}
\end{figure}

The diffusion model was then conditioned on GFS data to generate high-resolution forecasts for 11 am local time each day. In this experiment, we used N=90 samples from the diffusion model, as additional samples continued to increase prediction accuracy (Figure~\ref{fig:number_of_samples}), and samples are computationally inexpensive, taking only seconds to generate. Once again the mean of the diffusion samples had lower RMSE than both the coarse resolution forecast it was conditioned on (GFS), and the time-conditioned persistence model. 

\section{Conclusion}

Machine learning methods developed for super-resolution or video-prediction are well-suited for problems in weather and climate forecasting. We have proposed a method for using score-based diffusion models to super-resolve the output of numerical weather models and provide fully-probabilistic predictions of both historical and future weather patterns. In experiments, we evaluated this approach for day-ahead solar forecasts on the island of Oahu, and showed that the learned diffusion model exhibits three desirable qualities: (1) each sample is a realistic-looking image; (2) the samples are diverse; and (3) the mean of the samples is good point-estimate with lower RMSE than both coarse resolution numerical weather prediction alone and a high-resolution persistence baseline. 

In follow-up work, we are evaluating this approach for estimating the risks that would otherwise require running a large ensemble of numerical simulations. Sampling from a diffusion model takes only seconds, and can learn patterns in data that are inaccessible to coarse-grained simulations, so if the model is accurate enough it could provide a powerful tool for estimating the risk of rare events. Such a tool could be useful at multiple scales in weather and climate modeling. 

\section{Acknowledgements}
Support for this work comes from NSF \#OIA-2149133. Technical support and advanced computing resources from University of Hawaii Information Technology Services – Cyberinfrastructure, funded in part by the National Science Foundation CC* awards \#2201428 and \#2232862 are gratefully acknowledged.

\bibliography{nn,sadowski,atmo}
\bibliographystyle{icml2023}

\newpage
\appendix
\onecolumn
\section{Appendix}

\begin{figure}[htbp]
\centering
\hspace{0.5cm}
\includegraphics[scale=0.4]
{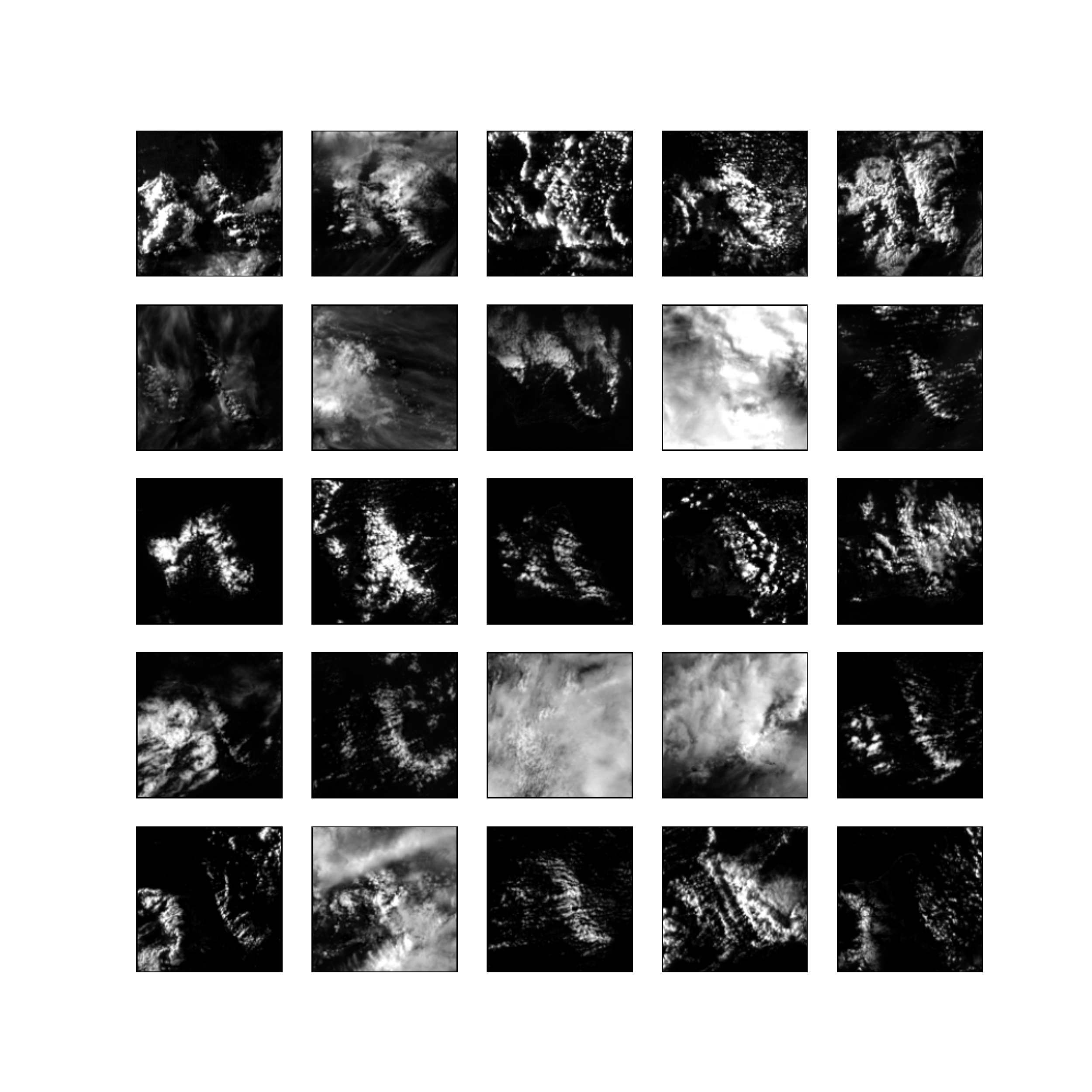}
\caption{Random data examples of the cloud coverage over the island of Oahu.}
\label{fig:gt_samples}
\end{figure}

\begin{table*}[ht!]
\caption{Hourly performance of our diffusion model on historical prediction.}
\vskip 0.15in
\begin{center}
\begin{small}
\begin{sc}
\begin{tabular}{lcccccc}
\toprule
Hour & Weather Model & Diffusion & Persistence & ERA5 & p-value\\
\midrule
6am & ERA5  & 0.246$\pm$ 0.077 & 0.258$\pm$ 0.084 & 0.343$\pm$ 0.150 & 0.002\\
7am & ERA5  & 0.237$\pm$ 0.072 & 0.248$\pm$ 0.075 & 0.343$\pm$ 0.157 & $<5\times 10^{-4}$\\
8am & ERA5  & 0.222$\pm$ 0.076 & 0.241$\pm$ 0.079 & 0.331$\pm$ 0.173 & $<5\times 10^{-4}$\\
9am & ERA5  & 0.202$\pm$ 0.071 & 0.219$\pm$ 0.082 & 0.336$\pm$ 0.190 & $<5\times 10^{-4}$\\
10am & ERA5  & 0.191$\pm$ 0.067 & 0.205$\pm$ 0.085 & 0.343$\pm$ 0.204 & $<5\times 10^{-4}$\\
11am & ERA5  & 0.188$\pm$ 0.067 & 0.202$\pm$ 0.089 & 0.357$\pm$ 0.201 & $<5\times 10^{-4}$\\
12am & ERA5  & 0.190$\pm$ 0.064 & 0.205$\pm$ 0.092 & 0.376$\pm$ 0.204 & $<5\times 10^{-4}$\\
1pm & ERA5  & 0.191$\pm$ 0.066 & 0.208$\pm$ 0.091 & 0.385$\pm$ 0.201 & $<5\times 10^{-4}$\\
2pm & ERA5  & 0.189$\pm$ 0.064 & 0.205$\pm$ 0.088 & 0.388$\pm$ 0.200 & $<5\times 10^{-4}$\\
3pm & ERA5  & 0.198$\pm$ 0.072 & 0.218$\pm$ 0.092 & 0.384$\pm$ 0.191 & $<5\times 10^{-4}$\\
4pm & ERA5  & 0.215$\pm$ 0.078 & 0.241$\pm$ 0.092 & 0.372$\pm$ 0.179 & $<5\times 10^{-4}$\\
5pm & ERA5  & 0.250$\pm$ 0.088 & 0.271$\pm$ 0.082 & 0.365$\pm$ 0.158 & $<5\times 10^{-4}$\\
6pm & ERA5  & 0.295$\pm$ 0.086 & 0.300$\pm$ 0.076 & 0.377$\pm$ 0.142 & 0.04\\
\midrule
Total & ERA5  & 0.207$\pm$ 0.075 & 0.223$\pm$ 0.090 & 0.362$\pm$ 0.190 & $<5\times 10^{-4}$\\
\bottomrule
\end{tabular}
\end{sc}
\end{small}
\end{center}
\vskip -0.1in
\label{tab:hourly_table}
\end{table*}

\begin{figure}[htbp]
\centering
\hspace{0.5cm}
\includegraphics[scale=0.4]
{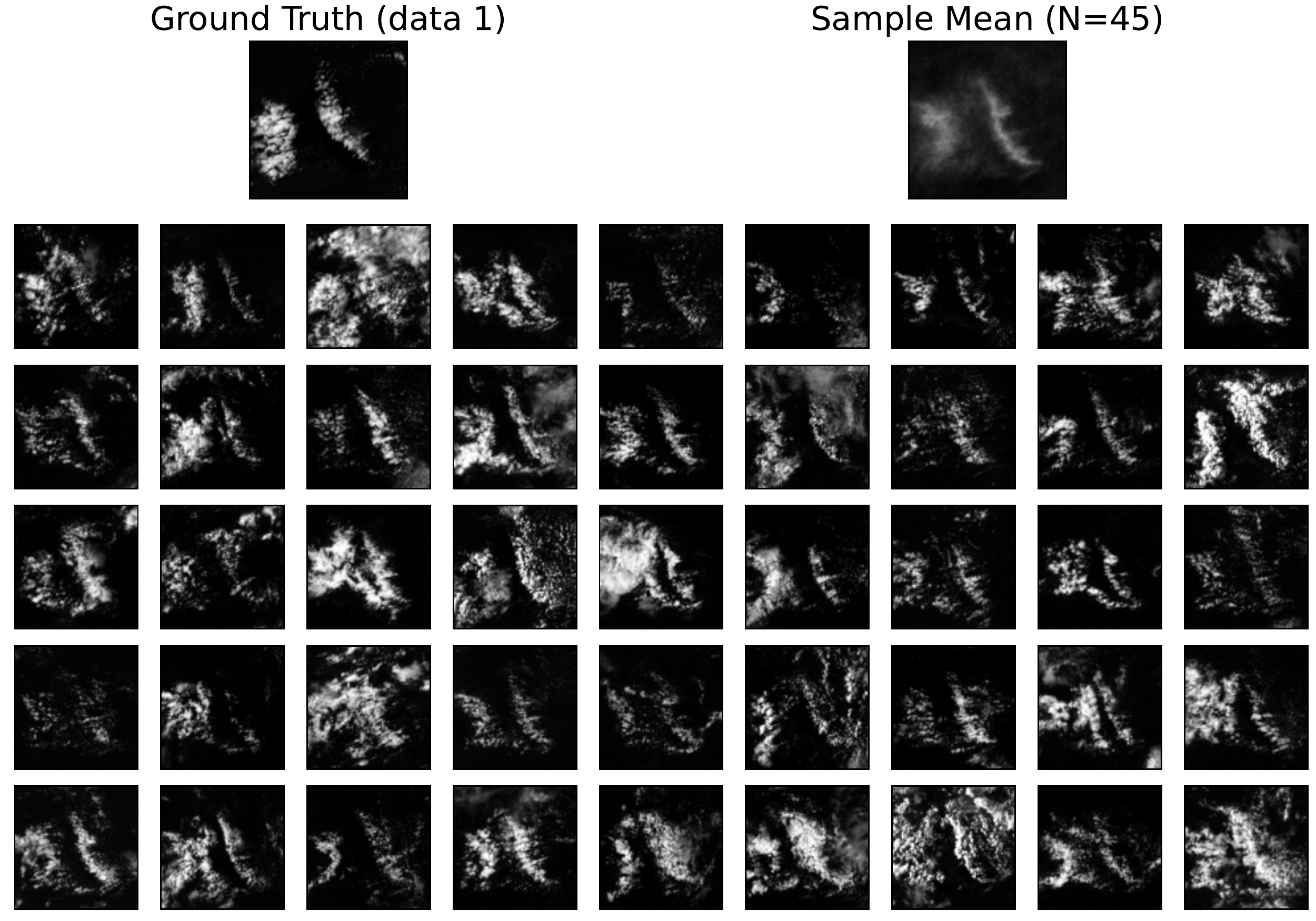}
\\\textcolor{lightgray}{\rule[1ex]{30em}{0.1pt}}\\
\includegraphics[scale=0.4]
{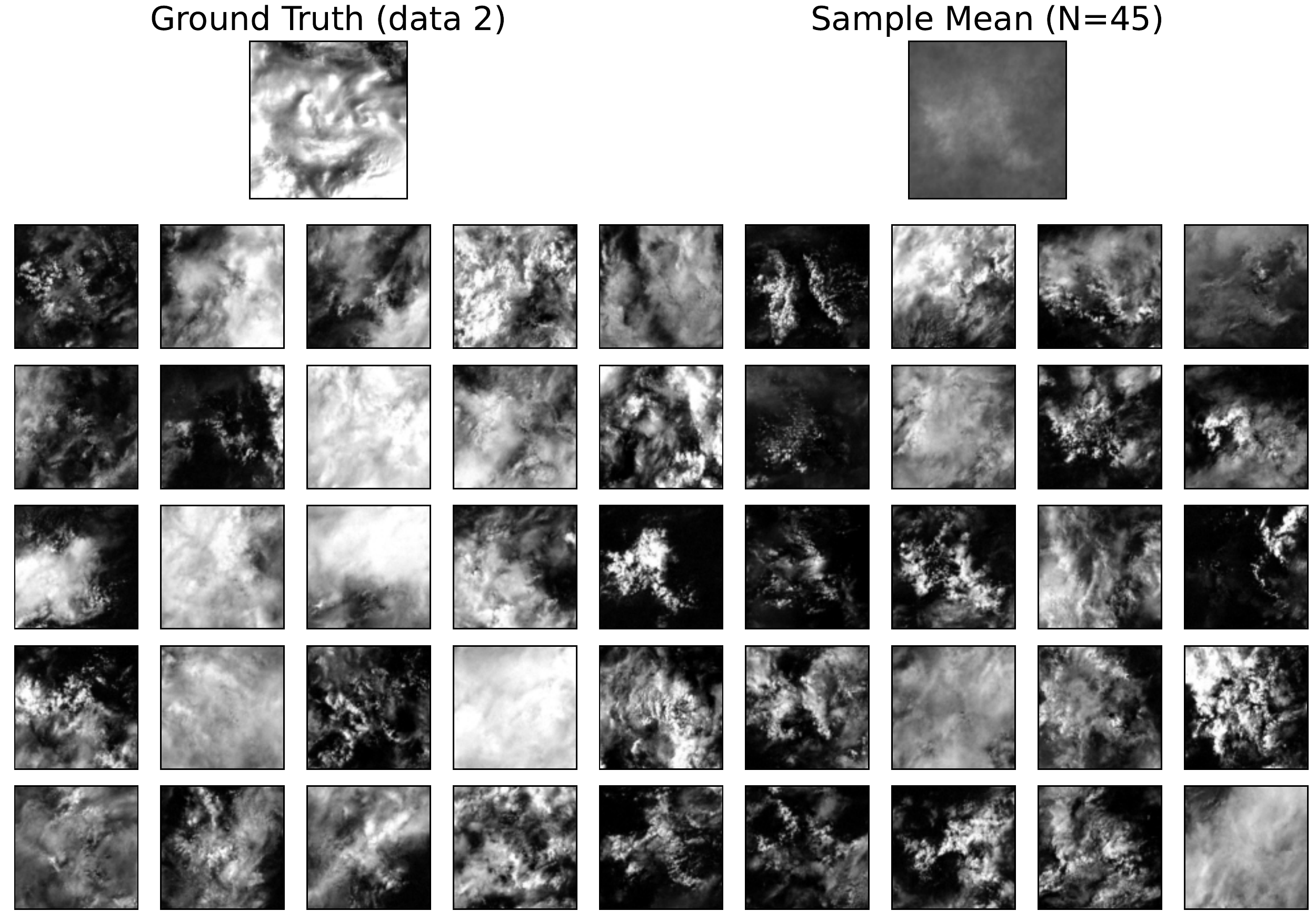}

\caption{All of the generated samples for historical prediction (data 1 and 2 in Fig. \ref{fig:samples_era5}).}
\label{fig:all_samples_era5}
\end{figure}


\end{document}